\pdfoutput=1
\documentclass[journal]{IEEEtran}

\usepackage{epsfig,epstopdf}
\usepackage{cite}
\usepackage{graphicx}
\usepackage[cmex10]{amsmath}

\usepackage{array}
\usepackage{amsfonts}

\usepackage{etoolbox}
\makeatletter
\patchcmd{\@makecaption}
{\scshape}
{}
{}
{}
\makeatletter
\patchcmd{\@makecaption}
{\\}
{.\ }
{}
{}
\makeatother
\def\tablename{Table}

\usepackage{color, colortbl}
\definecolor{LightCyan}{rgb}{0.89,1,1}
\definecolor{DarkCyan}{rgb}{0.5,1,1}

\begin{document}
%
\title{Robust Environmental Sound Recognition with Sparse Key-point Encoding and Efficient Multi-spike Learning}
%
%
%

\author{
	Qiang~Yu,~\IEEEmembership{Member,~IEEE,} 
	Yanli~Yao, 
	Longbiao~Wang,
	Huajin~Tang,~\IEEEmembership{Member,~IEEE,}
	Jianwu~Dang,~\IEEEmembership{Member,~IEEE,}
	Kay~Chen~Tan,~\IEEEmembership{Fellow,~IEEE}

\thanks{Q.~Yu, Y.~Yao, L.~Wang and J.~Dang are with Tianjin Key Laboratory of Cognitive Computing and Application, College of Intelligence and Computing, Tianjin University, Tianjin, China (e-mail: \{yuqiang, yaoyanli, longbiao\_wang\}@tju.edu.cn, jdang@jaist.ac.jp).}

\thanks{H.~Tang is with the College of Computer Science, Sichuan University, Chengdu, China (e-mail: htang@scu.edu.cn).}

\thanks{J.~Dang is also with Japan Advenced Institute of Science and Technology, Japan.}

\thanks{K.C.~Tan is with the Department of Computer Science, City University of Hong Kong, Hong Kong (e-mail: kaytan@cityu.edu.hk).}

\thanks{Manuscript is under review at ieee transactions.}

}

%
%

\markboth{}%
{Shell \MakeLowercase{\textit{et al.}}: Bare Demo of IEEEtran.cls for Journals}
%



\maketitle

\begin{abstract}
The capability for environmental sound recognition (ESR) can determine the fitness of individuals in a way to avoid dangers or pursue opportunities when critical sound events occur. It still remains mysterious about the fundamental principles of biological systems that result in such a remarkable ability. Additionally, the practical importance of ESR has attracted an increasing amount of research attention, but the chaotic and non-stationary difficulties continue to make it a challenging task. In this study, we propose a spike-based framework from a more brain-like perspective for the ESR task. Our framework is a unifying system with a consistent integration of three major functional parts which are sparse encoding, efficient learning and robust readout.
We first introduce a simple sparse encoding where key-points are used for feature representation, and demonstrate its generalization to both spike and non-spike based systems. Then, we evaluate the learning properties of different learning rules in details with our contributions being added for improvements. Our results highlight the advantages of the multi-spike learning, providing a selection reference for various spike-based developments. Finally, we combine the multi-spike readout with the other parts to form a system for ESR. Experimental results show that our framework performs the best as compared to other baseline approaches. In addition, we show that our spike-based framework has several advantageous characteristics including early decision making, small dataset acquiring and ongoing dynamic processing. Our framework is the first attempt to apply the multi-spike characteristic of nervous neurons to ESR. The outstanding performance of our approach would potentially contribute to draw more research efforts to push the boundaries of spike-based paradigm to a new horizon.


\end{abstract}

%
\IEEEpeerreviewmaketitle

\section{Introduction}

\IEEEPARstart{E}{nvironmental} sound recognition is an important ability of an
individual to quickly grasp useful information from ambient environment, the success of which can lead to prompt actions to be taken before potential opportunities or dangers, and thus determine the fitness of the individual. For example, a prey may realize the approach of a predator by the sounds of breaking twigs even without a vision. A successful recognition of such sounds often means its survival. Human and other animals are very good at recognizing environmental sounds. This extraordinary ability has inspired more efforts being devoted to endow artificial systems with a similar ability for environmental sound recognition (ESR) \cite{wang2006computational,cowling2003comparison,sharan2016overview,chachada2014environmental}, which can be also referred as automatic sound recognition (ASR).

ESR has attracted increasing attention in recent years from the field of acoustic signal processing \cite{o2008automatic,cowling2003comparison,sharan2016overview}, as well as neuroscience \cite{leaver2010cortical}. Similar to other well studied tasks such as speech or music recognition, ESR aims to recognize a specific sound automatically from the environment. Differently, the chaotic and unstructured difficulties residing in the sound signals make ESR a challenging and distinctive task. 
In addition, its practical importance has been reflected by a number of various newly applied developments or attempts including, but not limited to, bioacoustic monitoring \cite{weninger2011audio}, 
surveillance \cite{ntalampiras2009acoustic} and general machine hearing \cite{lyon2010machine}. Successful recognition of critical sounds like gunshots can send an early alarm to crowd and police, and thus help to save more lives and minimize losses. 
An ESR system can provide machines like a robot \cite{pineau2003towards} a cheap and advantageous way to understand the environment under poor visual conditions such as weak lighting or visual obstruction. Compared to vision-based processing, audio signals are relatively cheap to compute and store, which brings benefits of high efficiency and low-power consumption.
Both research challenges and advantages have motivated studies designed for ESR systems.

A general approach to pattern recognition tasks can be used to ESR. The approach typically contains three key steps \cite{sharan2016overview,Yu2013TNN} which are signal preprocessing, feature extraction and classification. These steps are tightly jointed to facilitate the functionality as a whole: signal preprocessing aims to prepare sound information for a better feature extraction which will then improve the performance of the classification.
In its primitive phases, ESR algorithms were simple reflections of speech and music processing paradigms \cite{cowling2003comparison,sharan2016overview,chachada2014environmental,huang2001spoken}, but divergence emerges as considerably non-stationary characteristics of environmental sounds are taken into account. The recognition performance of ESR systems largely depends on the choice of two essential components, i.e. feature(s) and classifier(s).
Research efforts have been made to them, as well as different combinations of methods from each. Different approaches can thus be categorized by methods being adopted for each component.

Features developed for speech processing are often used for ESR \cite{mitrovic2010features}. Statistical features are introduced to give descriptions of the sound signal in terms of psychoacoustic and physical properties such as power, pitch and zero-crossing rate, etc. These features are often only used as supplementary ones in ESR systems \cite{rabaoui2008using}. Cepstral features such as Mel-Frequency Ceptral Coefficients (MFCC) and spectrogram images are the most frequently used ones. The frame-based MFCC features are more favorable for modeling single sound sources but not for environmental sounds which typically contain a variety of sources \cite{chu2009environmental}. In addition, MFCC features are modeled from the overall spectrum which makes them vulnerable to noise \cite{dennis2013temporal}. On the other hand, spectrogram images are good at describing acoustic details of the sound from both the time and frequency domain \cite{dennis2011spectrogram}, but high dimensionality of the feature restricts its applicability \cite{sharan2016overview,chachada2014environmental}. There has been increasing number of advanced and sophisticated feature representations, such as stabilized auditory image \cite{lyon2010sound} and matching pursuit \cite{chu2009environmental}. Some other works \cite{dennis2013temporal,wu2018spiking} construct representations by utilizing additional feature extraction methods like self-organizing map (SOM). However, complexity of these feature representations is one of the major drawbacks. Recently, a simpler and more efficient feature representation method for sounds has been introduced by using local time-frequency information (LTF) \cite{xiao2018spike}. We will continue to contribute toward this method with simplicity, sparseness and flexibility bearing in mind.

Various classifiers have been successfully applied to ESR tasks in recent years.
The most commonly used classifiers \cite{kolozali2013automatic,chu2009environmental,lu2003content} include multi-layer perceptron (MLP), k-nearest neighbor (kNN), Gaussian mixture model (GMM) and support vector machines (SVMs). These classifiers continue to be used with modifications or a hybrid of classification algorithms \cite{sharan2016overview,chu2009environmental}, but they ignore the temporal information of sound signals. The hidden Markov model (HMM) was then applied to capture the temporal structure for a better performance \cite{dennis2011spectrogram}. However, HMM do not model explicitly the diverse temporal dependencies of environmental sounds \cite{dennis2013temporal}, leading further research towards a more complete modeling of the temporal structure.
In recent years, artificial neural networks (ANNs) with a class of techniques called deep learning have been thriving with a great success in various recognition tasks \cite{lecun2015deep}. Two of the most popular deep learning structures are deep neural network (DNN) and convolutional neural network (CNN), which have been successfully applied to ESR tasks very recently \cite{piczak2015environmental,zhang2015robust,ozer2018noise,mcloughlin2015robust}.
One of the major challenges of the aforementioned classifiers is the biological plausibility. Human brain is remarkably good at various cognitive tasks, including sound recognition, with extraordinary performance in terms of accuracy, efficiency, robustness and adaptivity, etc. How to transfer these advantageous abilities of the brain to artificial systems for solving ESR tasks motivates our study in this work.

\begin{figure}[!htb]
	\centering\includegraphics[width=0.48\textwidth]{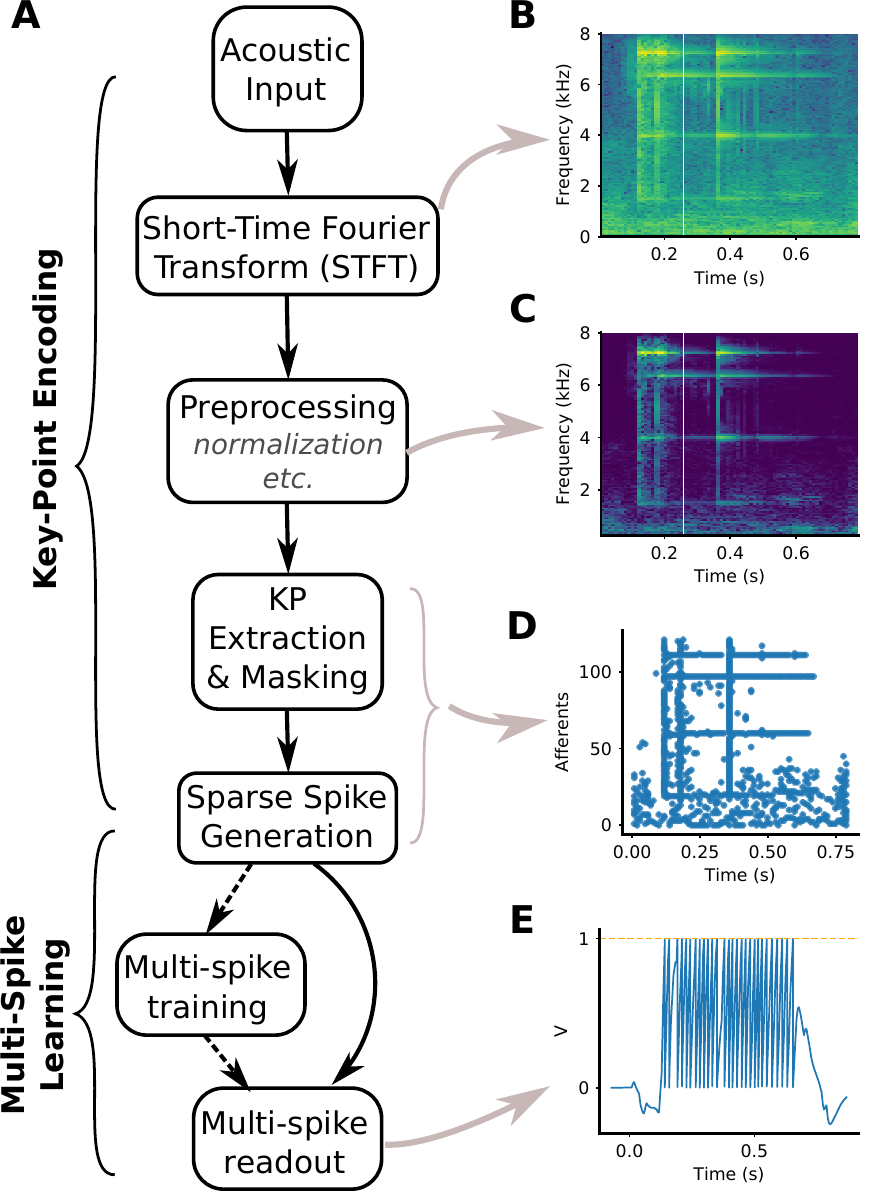}
	\caption{Overall framework of the multi-spike neural network for sound recognition.
	\textbf{A}, information processing structure of the system. \textbf{B}-\textbf{E},  demonstration of information in different systematic components: spectrogram images (\textbf{B} and \textbf{C}), spatiotemporal spike pattern (\textbf{D}) and dynamics of neuron's membrane potential (\textbf{E}).}
	\label{Fig:framework}
\end{figure}

Neurons in the brain use spikes, also called electrical pulses, to transmit information between each other \cite{kandel2000principles}. The discrete feature of spikes is believed to play an essential role in efficient computation, which has inspired a group of neuromorphic hardware implementations \cite{benjamin2014neurogrid,merolla2014million,yao2017face}. In spite of these hardware developments, how could spikes convey information still remains unclear. A sequence of spikes could encode information either with the total number of spikes or their precise spike timings, representing two of the most popular neural coding schemes, i.e. rate and temporal codes, respectively \cite{kandel2000principles,dayan2001theoretical,YuNCS,Panzeri10}. The rate code ignores the temporal structure of the spike train, making it highly robust with respect to interspike-interval noise \cite{stein2005neuronal,london2010sensitivity}, while the temporal code has a high information-carrying capacity as a result of making full use of the temporal structure \cite{Hopfield95,Richard98,Borst99}. Although an increasing number of experiments have been shown in various nervous systems \cite{london2010sensitivity,kandel2000principles,gollisch2008rapid,butts2007temporal} to support different codes, it is still arguable whether the rate or temporal code dominates information coding in the brain \cite{masuda2002bridging,gutig2014spike,yu2018spike}.

Different learning algorithms have been developed to better understand the underlying computing and processing principles of neural systems. One of the most widely studied rules is spike-timing-dependent plasticity (STDP) \cite{dan2004spike,song2000competitive} which instructs neurons to update their synaptic efficacies according to the temporal difference between afferent and efferent spikes. Dependence of temporal continuity hinders its development to act as an appropriate classifier \cite{gutig06,Yu2013TNN}.
The tempotron learning rule \cite{gutig06} is proposed to discriminate target and null patterns by firing a single spike or keeping silent. The firing-or-not behavior of the tempotron makes it an efficient learning rule, but constrains its ability to utilize the temporal structure of the neuron's output \cite{yu2018spike}.
Some other learning rules are proposed to train neurons to fire spikes at desired times \cite{bohte02spikeprop,ponulak10,florian2012chronotron,mohemmed2012span,YuQ2013PSD,memmesheimer2014}.
Although the temporal structure could be utilized by precise output spikes, designing an instructor signal with precise timings is challenging for both biological and artificial systems. Additionally, these learning rules are developed under the assumption of a temporal code, resulting that most of them cannot be generalized to a rate code \cite{yu2018spike,brette2015philosophy}. In \cite{gutig2016}, a novel type of learning rule is developed to train neurons to fire a desired number of spikes rather than precise timings. This multi-spike learning rule thus provide a new way to overcome limitations of the other ones. Improved modifications have been developed in \cite{yu2018spike,yu2018iconip}, along with detailed evaluations of different properties as well as theoretical proofs.

How to adopt the biologically plausible network, i.e. spiking neural network (SNN), to the ESR task demands more efforts. Previous related works \cite{dennis2013temporal,wu2018spiking,xiao2018spike} to this research problem have demonstrated the advantages of the spike-based approach. Following a general processing structure with SNN \cite{Yu2013TNN,yu2016spiking}, the framework normally consists of three functional parts, i.e. encoding, learning and readout. 
In \cite{dennis2013temporal,wu2018spiking}, the encoding depends on an SOM model, which could complicate the process and thus degrade computing efficiency. In addition, the biologically plausible implementation of this SOM model remains challenging, let alone a precise time reference to each segmentation frame for encoding spikes \cite{wu2018spiking}. In the learning part, the approaches of \cite{dennis2013temporal,wu2018spiking,xiao2018spike} are based on a binary-spike tempotron rule, which will limit neurons' capability to fully explore and exploit the temporal information over the presence of sounds. The readout in \cite{dennis2013temporal,wu2018spiking} relies on a voting scheme over the maximal potential. This means a recorder is required for tracking this maximum, and thus the efficiency and effectiveness of the readout is degraded.

In this work, we propose a spike-based framework (see Fig.~\ref{Fig:framework}) for the ESR task by combining a sparse key-point encoding and an efficient multi-spike learning. The significance of our major contributions can be highlighted in the following five aspects.

\begin{itemize}
	\item An integrated spike-based framework is developed for ESR. The event-driven scheme is employed in the overall system from encoding to learning and readout without taking advantages of other auxiliary traditional methods like SOM, making our system more consistent, efficient and biologically plausible. Compared to other non-spike-based approaches (which we refer as conventional ones), our system contributes to drive a paradigm shift in the processing way towards more human-like.
		
	\item A simplified key-point encoding frontend is proposed to convert sound signals into sparse spikes. The key-points are directly used without taking any extra steps of feature clustering. This simplification could be beneficial for low-power and on-line processing. Moreover, we show the effectiveness of our encoding by combining it with two of the most popular networks, i.e. CNN and DNN. Their performance is significantly improved, indicating the generalization of our encoding.
	
	\item We extend our previous multi-spike learning rule, namely threshold-driven plasticity (TDP) \cite{yu2018spike}, to solve the practical ESR task. A novel range training mechanism is developed to enhance the capability of the learning rule.
	Moreover, we examine its properties including efficiency, robustness and capability of learning inhomogeneous firing statistics. We are the first one, to the best of our knowledge, to make detailed comparisons of three representative learning rules. 
	
 	\item The proposed system is robust to noise under mismatched condition. Benchmark results highlight the significance of our approach. Further improvement can be obtained with a multi-condition training method.
	
	\item Our system is robust to severe non-stationary noise and is capable of processing ongoing dynamic environmental sound signals. This highlights the applicability of our proposed system.
	
\end{itemize}

The remainder of this paper is structured as follows. Section~\ref{sec:Methods} details our proposed approaches and methods being applied. Section~\ref{sec:experiments} then presents our experimental results, followed by discussions in Section~\ref{sec:discuss}. Finally, we conclude our work in Section~\ref{sec:Conclusion}.



\section{Methods}
\label{sec:Methods}
In this section, we will introduce the components and methods used in our framework (see Fig.~\ref{Fig:framework}). Firstly, we describe the proposed encoding frontend that converts sound signals into spikes. Then, the neuron model is described, followed by various learning rules including tempotron \cite{gutig06}, PSD \cite{YuQ2013PSD,yu2016spiking} and TDP \cite{yu2018spike}. Additionally, we present our new methods for improving the performance of the multi-spike learning rules. Finally, other task-related methods and approaches are detailed.

\subsection{Key-Point Encoding Frontend}

\begin{figure}[!htb]
	\centering\includegraphics[width=0.47\textwidth]{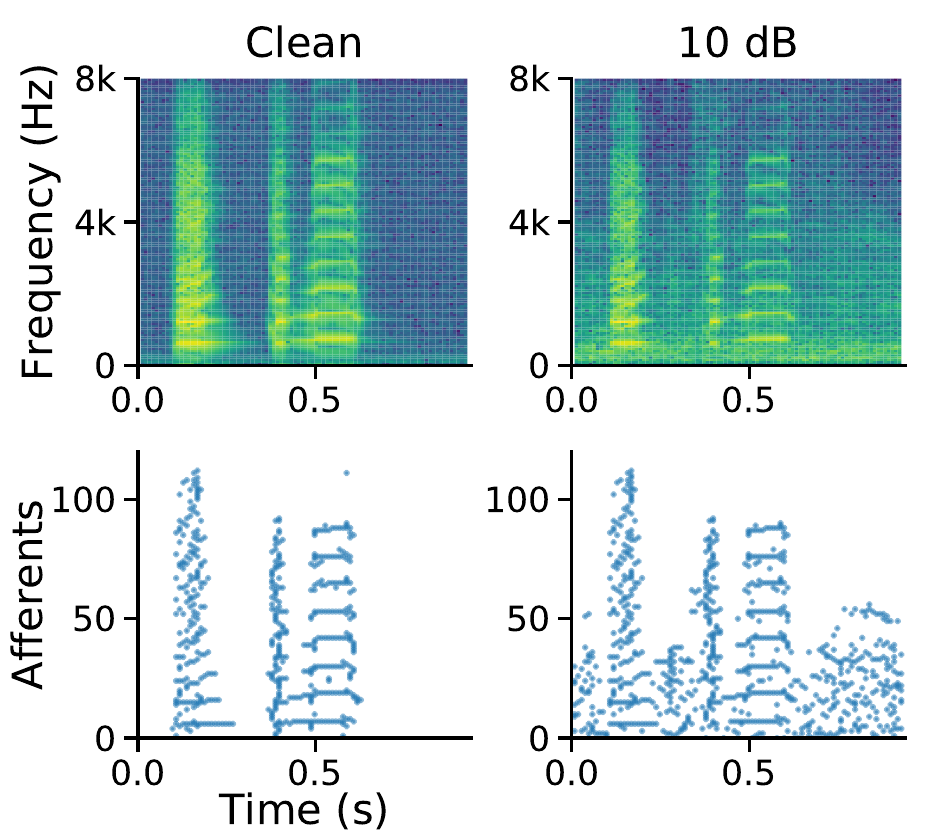}	
	\caption{Demonstration of key-point frontend encoding. A sound sample of `horn' is presented under conditions of clean and 10 dB of noise in the form of spectrogram (top row). The corresponding spike patterns encoded with our frontend are demonstrated in the bottom row. Each dot represents a spike from the corresponding afferent.}
	\label{Fig:kp}
\end{figure}

Biological evidence \cite{christopher1998optimizing,theunissen2000spectral,joris2004neural} gained from measurements of spectral-temporal response fields suggests that auditory neurons are sensitive to feature parameters of stimuli including local frequency bands, intensity, amplitude and frequency modulation, etc. The capability of neurons to capture local spectral-temporal features inspires the idea of utilizing key-points to represent sound signals \cite{dennis2013temporal,xiao2018spike}. Here, we present a more simpler and versatile encoding frontend by directly utilizing the key-points while keeping the processing steps to be as minimal as possible.

The detailed procedures of our encoding is presented in Fig.~\ref{Fig:framework}\textbf{A}. Sound signals are firstly converted into spectrograms by Short-Time Fourier Transform (STFT) with a window of 256 samples and a sliding step of 10 ms. The resulting spectrogram, $S(t,f)$, describes the power spectral density of the sound signal over both time and frequency dimensions (see Fig.~\ref{Fig:framework}\textbf{B}). Next, we perform a logarithm step to 
convert the spectrogram into a log scale through $log(S(t,f)+\epsilon)-log(\epsilon)$ with $\epsilon=10^{-5}$, followed by a normalization step. The resulting spectrogram (see Fig.~\ref{Fig:framework}\textbf{C}) which is still denoted as $S(t,f)$ for simplicity is further processed in the following key-point extraction steps.

The key-points are detected by localizing the sparse high-energy peaks in
the spectrogram. Such localizations are accomplished by searching local maxima across either time or frequency, as follows:
\begin{equation}
\label{eq:kp}
P(t, f) = \Bigg\{ S(t, f) \Big| S(t, f)=max \Big\{ \begin{array}{l}
S(t\pm d_t, f) \mbox{ or} \\
S(t, f\pm d_f)
\end{array} \Big \} \Bigg \}
\end{equation}
where $d_t (d_f)=[0,1,2,...,D_t(D_f)]$. $D_t$ and $D_f$ denote the region size for key-point detection. We set both of them to 4, which we found was big enough for a sparse representation, but small enough to extract important peaks.

In order to further enhance the sparseness of our encoding, we introduce two different masking schemes, namely the absolute-value and the relative-background masking. 
In the absolute-value scheme, those key-points are discarded if criterion of $P(t, f)<\beta_\mathrm{a}$ is satisfied. This means we only focus on significantly large key-points which are believed to contain important information. 
In the relative-background masking scheme, the key-point is dropped if the contrast between it and its surrounding background reaches the condition of
$P(t, f)*\beta_\mathrm{r}<mean\{S(t\pm d_t, f\pm d_f)\}$. $\beta_\mathrm{a}=0.15$ and $\beta_\mathrm{r}=0.85$ are the two hyper-parameters that control the level of reduction in the number of key-points.

The extracted key-points contain both spectral and temporal information, which we found is sufficient enough to form a spatiotemporal spike pattern by directly mapping each key-point to a spike. As can be seen from Fig.~\ref{Fig:kp}, the resulting spike pattern is capable of giving a `glimpse' of the sound signal with a sparse and robust representation. The advantageous properties of our encoding can be beneficial to learning algorithms, which we will show later.

\subsection{Neuron Model}

In this paper, we use the current-based leaky integrate-and-fire neuron model due to its simplicity and analytical tractability \cite{gutig2016,yu2018spike}.
The neuron continuously integrates afferent spikes into its membrane potential,
and generates output spikes whenever a firing condition is reached. Each afferent spike will result in a post-synaptic potential (PSP), whose peak value is controlled by the synaptic efficacy, $w$. The shape of PSPs is determined by the kernel defined as
\begin{equation}
K(t-t_i^j) = V_0\left[\exp{\left(-\frac{t-t_i^j}{\tau_\mathrm{m}}\right)}-\exp{\left(-\frac{t-t_i^j}{\tau_\mathrm{s}}\right)}\right]  
\end{equation}
where $V_0$ is a constant parameter that normalizes the peak of the kernel to unity, and $t_i^j$ denotes the time of the j-th spike from the i-th neuron. $\tau_\mathrm{m}$ and $\tau_\mathrm{s}$ represent time constants of the membrane potential and the synaptic currents, respectively.

\begin{figure}[!htb]
	\centering\includegraphics[width=0.47\textwidth]{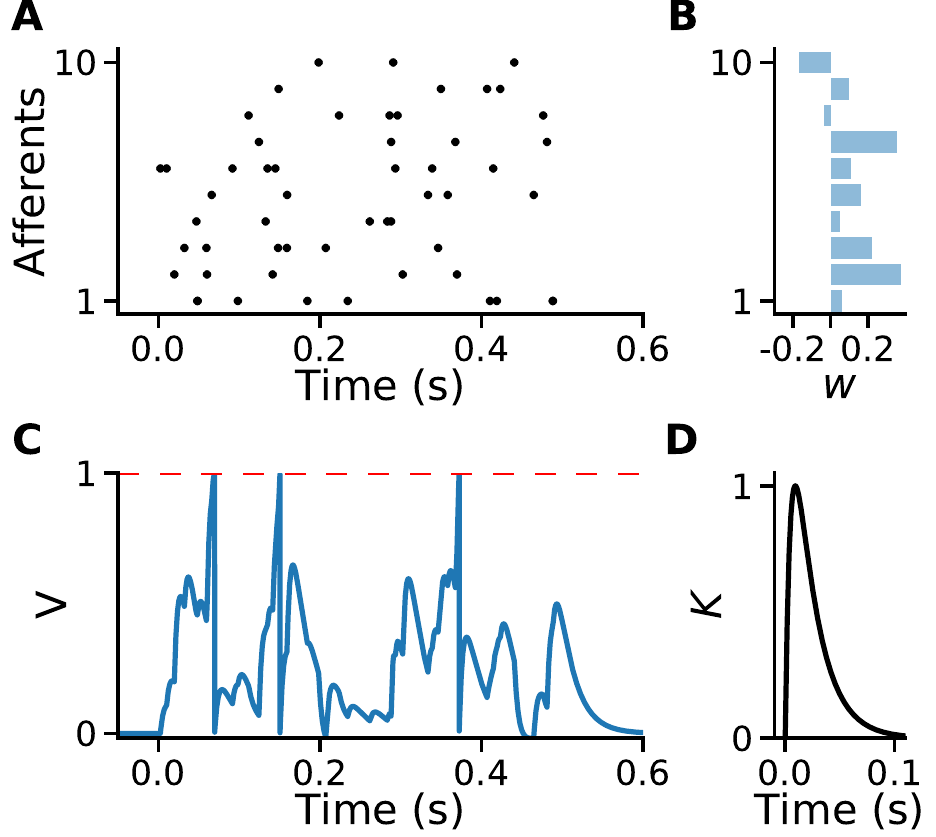}	
	\caption{Dynamics of spiking neuron model. \textbf{A}, the input spatiotemporal spike pattern where each dot denotes a spike. Each afferent fires a certain number of spikes across the time window. \textbf{B}, the synaptic weights of the corresponding afferents. \textbf{C}, membrane potential dynamics of the neuron in response to the pattern in \textbf{A}. The red dashed line denotes the firing threshold. \textbf{D}, normalized kernel of post-synaptic potential.}
	\label{Fig:NeuronDyms}
\end{figure}

The evolving dynamics of the neuron with $N$ synaptic afferents is described as
\begin{equation}
\label{Eq:neuron}
V(t) = \sum_{i=1}^N w_i\sum_{t_i^j<t} K(t-t_i^j) - \vartheta \sum_{t_\mathrm{s}^j<t} \exp{\left(-\frac{t-t_\mathrm{s}^j}{\tau_\mathrm{m}}\right)} 
\end{equation}
where $\vartheta$ denotes the firing threshold and $t_\mathrm{s}^j$ represents the time of the $j$-th output spike.

As can be seen from Fig.~\ref{Fig:NeuronDyms}, each afferent spike will result in a change in the neuron's membrane potential. The neuron continuously integrates afferent spikes in an event-driven manner. The mechanism of the event-driven computation is advantageous in both efficiency and speed \cite{yu2018spike}, and thus is adopted in our study. In the absence of an input spike, neuron's membrane potential will gradually decay to the rest level. Whenever the membrane potential crosses neuron's firing threshold, an output spike is elicited, followed by a reset dynamics.

\subsection{Learning Rules}

Various learning rules have been introduced to train neurons to learn spikes. These learning rules can be categorized according to neuron's output response. With supervised temporal learning rules, neurons can be trained to have an efficient binary response (e.g. spike or not) or multiple output spikes where either their precise timing or the total spike number matters. In our study, we select tempotron \cite{gutig06}, PSD \cite{YuQ2013PSD} and TDP \cite{yu2018spike} as representatives of different types. Our new contributions to these learning rules are provided after descriptions about them.

\subsubsection{The tempotron rule}

Different from a multi-spike neuron model as described in Eq.~\ref{Eq:neuron}, the tempotron can only fire a single spike due to the constraint of a shunting mechanism \cite{gutig06}. Neuron is trained to elicit a single spike in response to a target pattern ($P^+$) and to keep silent to a null one ($P^-$). In this rule, a gradient descent method is applied to minimize the cost defined by the distance between the neuron's maximal potential and its firing threshold, leading to the follow:
\begin{equation}
\Delta w_i=\left\{\begin{array}{rl}
\lambda\sum_{t_i<t_{\mathrm{max}}}K(t_{\mathrm{max}}-t_i),  &\mbox{ if $P^+$ error;}\\
-\lambda\sum_{t_i<t_{\mathrm{max}}}K(t_{\mathrm{max}}-t_i), &\mbox{ if $P^-$ error;}\\
0, &\mbox{ otherwise.}
\end{array}
\label{eq:tempotronRule}
\right.
\end{equation}
where $t_{\mathrm{max}}$ denotes the time at the maximal potential and $\lambda$ represents the learning rate.

Decision performance of the tempotron rule can be improved by incorporating other mechanisms such as grouping \cite{Yu2013TNN} and voting with maximal potential \cite{dennis2013temporal}. The mechanism of maximal potential decision is applied for the tempotron rule in our study.

\subsubsection{The PSD rule}

The PSD rule is proposed to train neurons to fire at desired spike times in response to input spatiotemporal spike patterns, such that the temporal domain of the output can be potentially utilized for information transmission as well as multi-category classification \cite{YuQ2013PSD,yu2016spiking}. The learning rule is implemented to minimize the difference between the actual ($t_\mathrm{o}$) and the desired ($t_\mathrm{d}$) output spike times, and the learning rule is thus given as:
\begin{align}
\label{Eq:TrialL}
\Delta w_i 
&=\lambda \Bigg [ \sum_g\sum_fK(t_\mathrm{d}^g-t_i^f)H(t_\mathrm{d}^g-t_i^f) \\ \nonumber
&- \sum_h\sum_fK(t_\mathrm{o}^h-t_i^f)H(t_\mathrm{o}^h-t_i^f) \Bigg ]
\end{align}
where $H(\cdot)$ represents the Heaviside function.

According to the PSD rule, a long-term potentiation (LTP) will occur to increase the synaptic weights when the neuron fails to fire at a desired time, while a long-term depression (LTD) will decrease the weights when the neuron erroneously elicits an output spike. The distance between two spike trains can be measured by
\begin{equation}
Dist = \frac{1}{\tau}\int_0^\infty [f(t)-g(t)]^2dt
\label{Eq:Dist}
\end{equation}
where $f(t)$ and $g(t)$ are filtered signals of the two spike trains.
This distance metric can be used in both training and evaluation, but the choice of a critical value for termination in the training could be difficult. In this paper, we introduce a much simpler and efficient approach, i.e. the coincidence metric, to measure the distance. We introduce a margin parameter $\zeta$ to control the precision of the coincidence detection. We will treat the output spike time as a correct one if it satisfies the condition of $t_\mathrm{d}-\zeta \leq t_\mathrm{o} \leq t_\mathrm{d}+\zeta$. This margin parameter can facilitate the learning.

\subsubsection{The TDP rule}

\begin{figure}[!htb]
	\centering\includegraphics[width=0.47\textwidth]{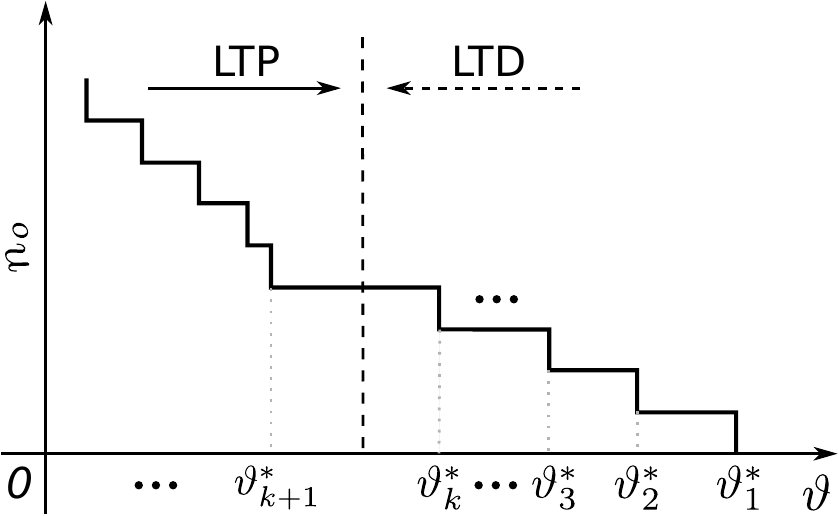}
	\caption{Demonstration of spike-threshold-surface (STS). Neuron's output spike number, $n_\mathrm{o}$, can be determined by the position of its firing threshold and the critical values $\vartheta^*_k$.}
	\label{Fig:sts}
\end{figure}

Recently, a new family of learning rules are proposed to train neurons to fire a certain number of spikes instead of explicitly instructing their precise timings \cite{gutig2016,yu2018spike}. These learning rules are superior to others for making decision and exploring temporal features from the signals. We adopt the TDP rule in this paper due to its efficiency and simplicity. The learning rule is developed based on the property of the multi-spike neuron, namely spike-threshold-surface (STS). Neuron's actual output spike number can often be determined by the position of its firing threshold in STS. Therefore, modifications of the critical threshold values can result in a desired spike number. The TDP learning rule is given as
\begin{equation}
\label{Eq:learning}
\Delta w =
\begin{cases}
-\lambda \frac{d\vartheta^*_{n_o}}{d w}       & \quad \text{if } n_\mathrm{o}>n_\mathrm{d} \\
\lambda \frac{d\vartheta^*_{n_o+1}}{d w}  & \quad \text{if } n_\mathrm{o}<n_\mathrm{d}\\
\end{cases}
\end{equation}
where $d\vartheta^*_{k}/d w$ represents the directive evaluation of critical threshold values with respect to synaptic weights (the details can be found in \cite{yu2018spike}). The basic idea of this learning rule (see Fig.~\ref{Fig:sts}) is to increase (decrease) the critical values that are smaller (greater) than $\vartheta$ with an LTP (LTD) process if the neuron fails to fire a desired number of spikes. The learning stops until the neuron's firing threshold falls into a desired region.

This learning mechanism makes the TDP rule capable of learning both rate- and temporal-based patterns \cite{yu2018spike}, which would be advantageous if the temporal structure of external stimuli is unknown.
In order to further enhance the applicability of the TDP rule, we develop a range training mechanism in this paper. Instead of using a specific desired spike number, the learning is stopped if neuron's output spike number falls into a desired range. In our sound recognition task, we train neurons to fire at least 20 spikes in response to their target categories.

\subsection{Deep Learning Networks}

CNN and DNN, as two of the most popular networks in deep learning \cite{lecun2015deep}, are also applied in this study to benchmark our proposed approaches.

A CNN typically consists of input, convolutional, pooling, normalization, fully connected and output layers.
Since CNN favors input images with fixed dimensions, we extend spectrograms to the longest duration of all sound signals by employing zero padding to the end. 
We set our CNN architecture to 32C3@127$\times$211-64C3@63$\times$105-128C3@31$\times$52-256C3@15$\times$25-F64-F10. 
It consists of 5 learning layers, including 4 convolutional and one fully connected layer. All learning layers use the non-linearity rectified linear units (ReLU) as the activation function, and batch normalization is applied to avoid over-fitting. In addition, we use 2$\times$2 strides in all learning layers except for the first convolutional one. The CNN network is trained using Adam optimizer with a learning rate of 0.0001.

DNN is a feed-forward artificial neural network, which consists of more than one layer of hidden units between the inputs and outputs. We construct a 4-layer DNN of the form 256-180-64-32 with the output layer in a one-of-N (i.e. N categories) configuration. We flatten the spectrograms to one-dimensional vectors which serve as the inputs of DNN. We adopt ReLU activation except for the input and output layers. Again, the Adam optimizer with a learning rate of 0.0001 is adopted.

\subsection{Sound Database}
Following the setups in \cite{dennis2013temporal,xiao2018spike,wu2018spiking}, we choose the same following ten sound classes for a fair comparison from the Real World Computing Partnership (RWCP) \cite{nakamura2000acoustical}: whistle1, ring, phone4, metal15, kara, horn, cymbals,
buzzer, bottle1 and bells5. 
To standardize the selection, we choose the first 80 files from each class to
form our experimental dataset. 
In each experimental run, we randomly select half files
of each class as the training set, and leave the rest as testing. The "Speech Babble" noise
environment is obtained from NOISEX’92 database \cite{varga1993assessment} for evaluating
the robustness of the sound recognition. The performance of different approaches is evaluated in both clean
environment and noisy cases with signal-to-noise ratio (SNR) of 20, 10, 0 and -5 dB. 
The performance is then averaged over 10 independent runs.

\section{Experimental Results}
\label{sec:experiments}

In this section, we first examine properties of different learning rules. To be specific, we concern the properties of learning efficiency and multi-category classification. Additionally, we also show the capability of the multi-spike learning rule for processing a more challenging task of inhomogeneous firing. Then, we present the performance of our proposed framework for sound recognition. Detailed examinations on various learning properties of the system are given accordingly. Finally, we show the outstanding performance of our system for processing ongoing dynamic environmental signals.

\subsection{Learning Efficiency of Multi-Spike Rules}

In this experiment, we evaluate the learning efficiency of different multi-spike learning rules, including PSD \cite{YuQ2013PSD}, MST \cite{gutig2016} and TDP \cite{yu2018spike}. These learning rules can be used to train neurons to fire a desired number of spikes. Different from the others, PSD rule requires precise spike timings in the supervisor signal. In order to relax this constraint, we employ margin parameters $\zeta$ of 5 and 10 ms to the PSD rule.

\begin{figure}[!htb]
	\centering\includegraphics[width=0.47\textwidth]{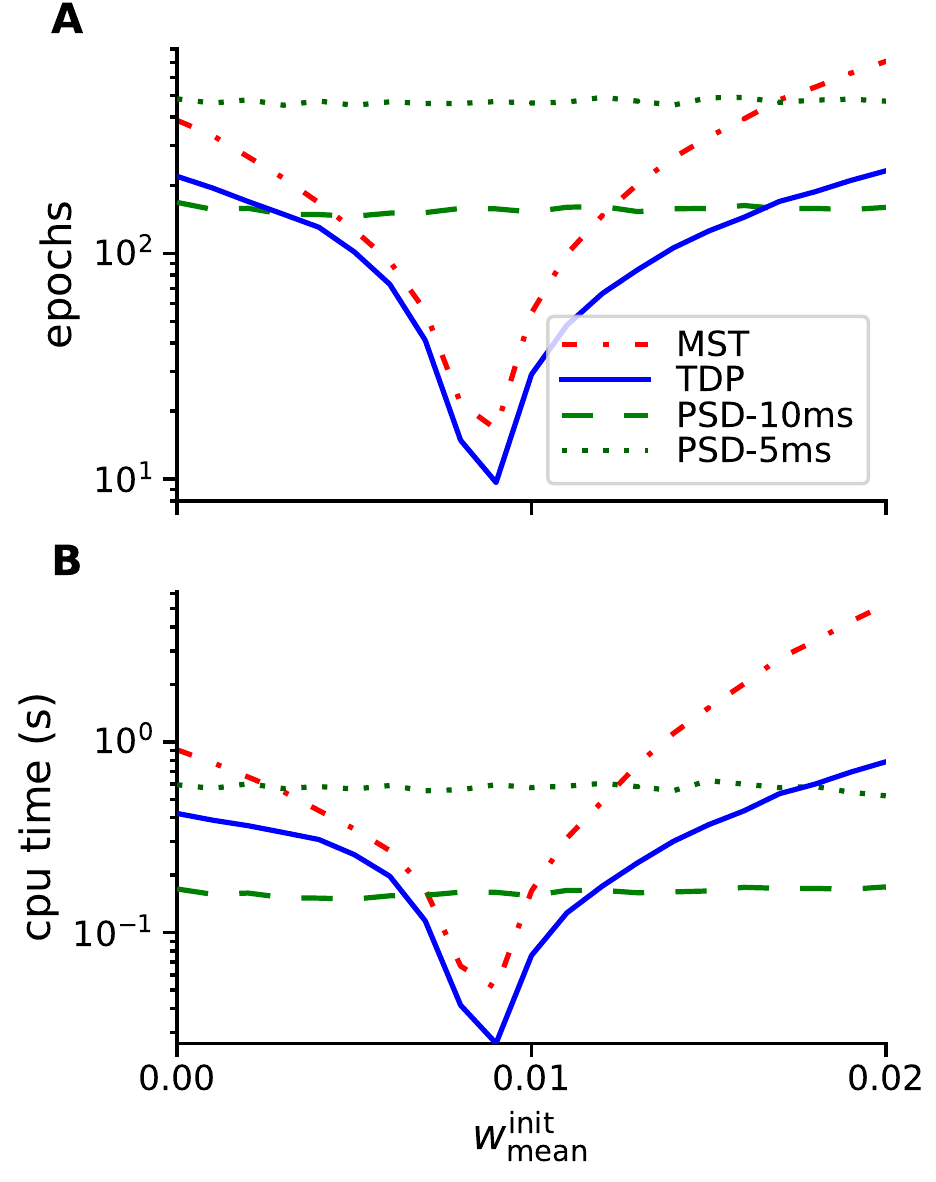}
	\caption{Efficiency of multi-spike learning rules including MST, TDP and PSD. \textbf{A} and \textbf{B} show the learning epochs and the corresponding cpu running time, respectively. Experiment was performed on a platform of Intel E5-2620@2.10GHz. All the learning rules are used to train neurons to fire 20 spikes. In the PSD rule, the desired times of these 20 spikes are constructed by evenly distributing them over the time window. Margin parameters of 5 and 10 ms are applied in the PSD rule. Results were averaged over 100 runs.}
	\label{Fig:efficiency}
\end{figure}

The input spike patterns are generated over a time window of 
T = 1.0 s with each afferent neuron firing at a Poisson rate of 8 Hz over T. 
Similar to \cite{yu2018spike}, other parameters are set as: $N=500$, $\tau_\mathrm{m}=20$ ms, $\tau_\mathrm{s}=5$ ms and $\lambda=10^{-4}$. Neurons are trained to elicit 20 spikes with different learning rules under different initial weight setups. The synaptic weights are initialized according to a Gaussian distribution where we keep the standard deviation of 0.01 fixed while change the mean value for different evaluations.

As can be seen from Fig.~\ref{Fig:efficiency}, the learning speeds of both MST and TDP rules change with different initial mean weight due to the incremental updating characteristics of the learning. The current training method of the MST and TDP rules is implemented in a way to increase or decrease one spike a time. Neuron's output spike number normally increases with bigger mean values of the initial weights \cite{yu2018spike}. These are the reasons why the speeds of both MST and TDP increase first and then decrease with increasing mean initial weights. Notably, TDP always outperforms MST in terms of efficiency. Different from the other two, the learning speed of PSD barely changes with different initial conditions. This is because PSD employs a form of batch updating where neurons are instructed by all desired spikes together during learning. These desired spikes are independent of the initial weight setups, resulting in a roughly steady learning speed. The learning speed of PSD can be increased by further relaxing the margin parameter $\zeta$ (e.g. from 5 ms to 10 ms). Although the margin scheme facilitates the learning, the precise spike timings of the instructor are still required.

\subsection{Learning to Classify Spatiotemporal Spike Patterns}

In this experiment, we study the ability of different rules on discriminating spatiotemporal spike patterns of different categories. The neuron parameters are the same as the previous experiment except that the mean and the standard deviation of initial weights are set as 0 and 0.001, receptively. Similar to the experimental setups in \cite{florian2012chronotron,mohemmed2012span,YuQ2013PSD}, we design a 3-category classification task and construct one template spike pattern for each category. Every template is randomly generated and then fixed after generation. Each afferent has a firing rate of 2 Hz over a time window of 0.5 s. Spike patterns of each category are instantiated by adding two types of noises to the template pattern. The first type is jitter noise: each spike of the pattern is jittered by a random Gaussian noise with zero mean and standard deviation of $\sigma_\mathrm{jit}$. The other type is deletion noise: each spike would be randomly deleted with a probability of $p_\mathrm{del}$. We use $\sigma_\mathrm{jit}=2$ ms and $p_\mathrm{del}=0.1$ to train neurons for the corresponding noise type, followed by evaluations over a broader range of noise. Each category is assigned to be the target of one learning neuron.

Two different readout schemes are applied: the absolute (`abs') and the winner-take-all (`wta') methods. In the `abs' method, the neuron with exactly the same output response as a predefined critical one will represent the prediction, while the one with the maximal response among all the output neurons will dominate the prediction in the `wta' method. We set different configurations for different learning rules. In the tempotron rule, we apply the binary response of spike or not for `abs', while the maximal potential is used in `wta'. In the PSD rule, neurons are trained to have 4 evenly distributed spikes and none for the target and null categories, respectively. A critical spike number of 2 is used in `abs', while the spike distance measurement of Eq.~\ref{Eq:Dist} is adopted in `wta'. A margin of 10 ms is also applied in the PSD. For the TDP rule, neurons are trained to elicit at least 20 spikes for target categories and keep silent for null ones. The critical spike number of 10 is used in `abs', while `wta' searches for the neuron with the maximal output spike number. In order to facilitate the learning, a momentum scheme \cite{gutig06,yu2018spike} with $\mu=0.9$ is also applied to all rules.

\begin{figure}[!htb]
	\centering\includegraphics[width=0.48\textwidth]{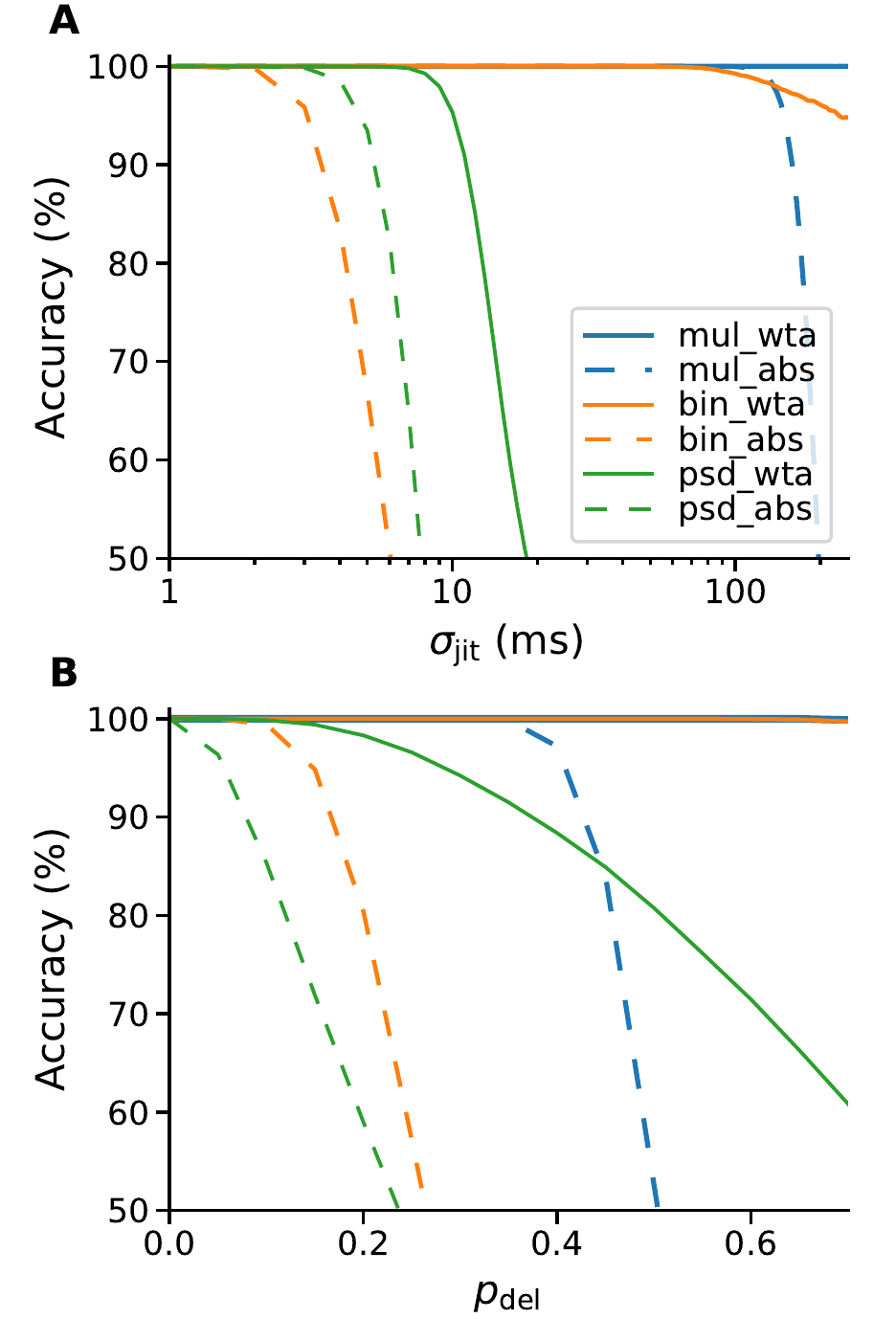}
	\caption{Robustness of various spike learning rules on spatiotemporal spike pattern classification. \textbf{A} and \textbf{B} show the classification accuracy against noises of spike jitter $\sigma_\mathrm{jit}$ and spike deletion $p_\mathrm{del}$, respectively. The examined learning rules are denoted as: `mul' (for TDP rule), `bin' (tempotron) and `psd' (PSD). These rules are combined with two readout schemes, i.e. `wta' and `abs', resulting in 6 combinations. Data were averaged over 100 runs.}
	\label{Fig:robustness}
\end{figure}

As can be seen from Fig.~\ref{Fig:robustness}, all the rules with `wta' scheme perform better than their counterparts with `abs'. This is because the competing policy in `wta' can help the system to identify the most likely representation by comparing all outputs. The TDP rule is the best as compared to other learning rules under both `wta' and `abs'. The performance of the tempotron rule is significantly improved by the maximal potential decision as compared to fire-or-not. This is because useful information can be integrated to subtle changes on the membrane potential which will be difficult to capture with a binary-spike response, but it could be reflected in the maximal potential to some extend. The PSD rule performs relatively worse than the others. This is because the desired spike timings would hardly be an optimal choice for a given task, and it is difficult to find a such one. In a sound recognition task, the appearance of a stimulus can be arbitrary, let alone to train a neuron to fire at desired times. Therefore, we only evaluate the performance of the tempotron and TDP in the following sound recognition tasks. Additionally, the `wta' scheme will be adopted due to its superior performance.

\subsection{Learning to Process Inhomogeneous Firing Statistics}
\begin{figure}[!htb]
	\centering\includegraphics[width=0.49\textwidth]{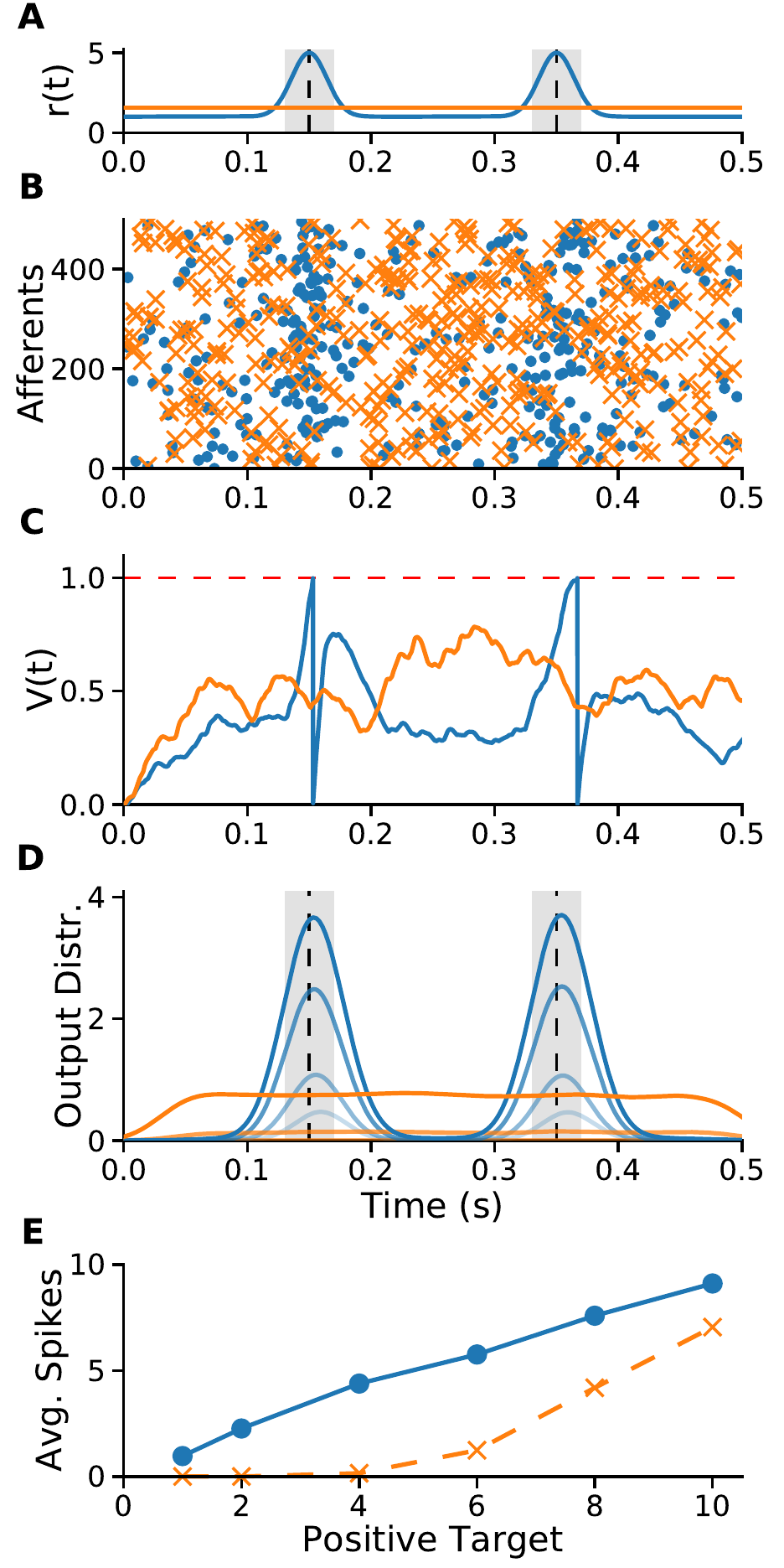}
	\caption{Learning performance of multi-spike rule on inhomogeneous firing statistics. \textbf{A}, firing rate $r(t)$ of two rate patterns as a function of time. The blue and orange colors represent the inhomogeneous (the target class) and homogeneous (the null class) firing cases, respectively. The vertical dashed lines depict the center of peaks, and shaded areas show the regions with a width to the left and right from the center. \textbf{B}, spike pattern instantiation examples of the corresponding firing rates. Each marker denotes a spike. \textbf{C}, voltage trace demonstration of a neuron that is trained to have 2 spikes for the target class and none for the null. \textbf{D}, illustration of output spike distribution over time in response to an input pattern. Color intensity from light to dark indicates a desired target spike number of 1, 2, 6 and 10. The curves of 1 and 2 for the null class (orange) are not clearly visualized since they are presented at zero. \textbf{E}, average output spike number versus the desired one used for the target class. \textbf{D} and \textbf{E} were obtained from 500 runs.}
	\label{Fig:rt}
\end{figure}

The nervous neurons general receive inputs of non-homogeneous statistics, i.e. time-varying firing rates. Thanks to the capability of the TDP rule to process both rate and temporal based patterns \cite{yu2018spike}, we will further our study to examine its capabilities of learning inhomogeneous firing statistics in this experiment.

We design two firing rate classes, i.e. inhomogeneous and homogeneous. The inhomogeneous firing rate is time-varying while it is constant for the homogeneous class. For the inhomogeneous rate, we add a form of $4*\exp^{-(\frac{t-c}{b})^2}$ to a baseline firing of 1 Hz, where $c$ and $b$ denotes the center and width, respectively. We choose two centers at 150 and 350 ms, and set the width to 20 ms. In order to remove statistical difference on the mean firing over the whole time window of 0.5 s, we set the homogeneous rate to a level such that the integrals of both firing statistics are the same. The two resulted classes of firing rate are shown in Fig.~\ref{Fig:rt}\textbf{A}. Spike patterns of each class are generated by instantiating spikes according to the instantaneous firing rate determined by $r(t)$. Different from the previous spatiotemporal task, we do not keep any generated spike patterns fixed, but always generate new ones according the corresponding $r(t)$. Fig.~\ref{Fig:rt}\textbf{B} demonstrates examples of two generated spike patterns. In this experiment, we use the TDP rule to train neuron to fire at least $n_\mathrm{d}$ spikes in response to the inhomogeneous patterns while keep silent to the homogeneous ones. 

Fig.~\ref{Fig:rt}\textbf{C} shows the learning results of a neuron with $n_\mathrm{d}$ being set to 2. The neuron can successfully elicit 2 spikes in response to a target pattern while keep silent as expected to a null one. Notably, the two output spikes occur around the peak centers with a nearly equal possibility, indicating a successful detection of the discriminative information. In order to have a detailed examination on this capability, we run the experiments with different $n_\mathrm{d}$ values. Fig.~\ref{Fig:rt}\textbf{D} and Fig.~\ref{Fig:rt}\textbf{E} show the output distribution and the total average output spike number in response to a pattern of different classes, respectively.
The output spike number for the target class increases with increasing $n_\mathrm{d}$, while it keeps around zero for the null class first at low $n_\mathrm{d}$ values (e.g. 4) and then starts to increase. Although it seems that the increasing output spike number of both classes makes the discrimination difficult, the distribution of these spikes over time can facilitate the classification (Fig.~\ref{Fig:rt}\textbf{D}). Neurons can always discover useful information by eliciting spikes around interesting areas, and downstream neurons can be added to further utilize these spikes.

\subsection{Environmental Sound Recognition}

In this experiment, we examine the capabilities of our framework on the task of sound recognition.
We set neuron's parameters as $\tau_\mathrm{m}=40$ ms and $\tau_\mathrm{s}=10$ ms. Synaptic weights are initialized with a mean of 0 and standard deviation of 0.01. Recognition performance is evaluated under clean and different noise levels of 20, 10, 0 and -5 dB.

\begin{figure}[!htb]
	\centering\includegraphics[width=0.48\textwidth]{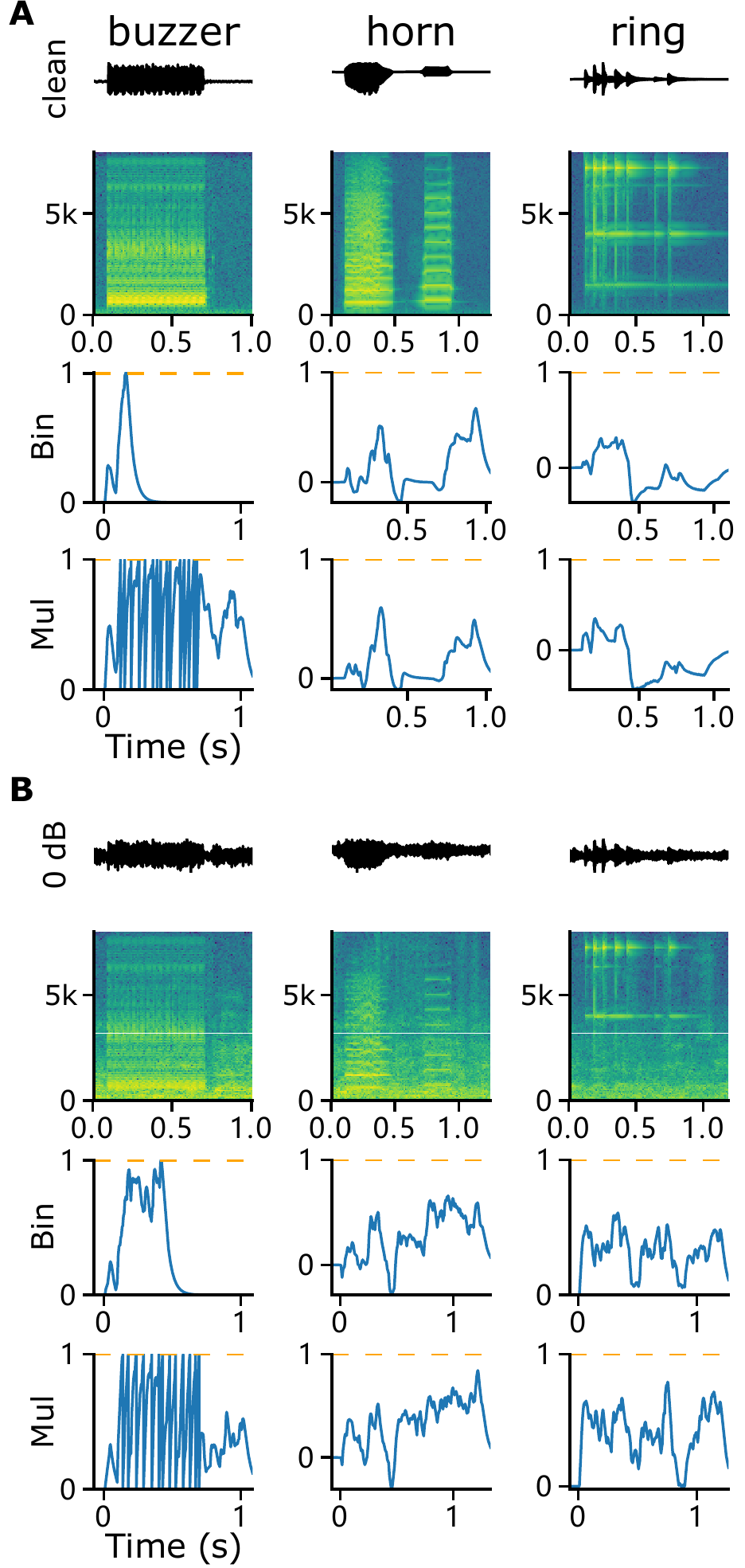}
	\caption{Demonstration of neurons' response to different sound samples under clean (\textbf{A}) and SNR of 0 dB background noise (\textbf{B}). The rows of each panel (from top to bottom) show: sound wave signals, spectrograms, voltage traces of a neuron with the tempotron rule (denoted as `Bin') in response to the corresponding sound samples, and voltages of a neuron with the TDP rule (represented by `Mul'). The target class of the demonstrated neurons is `buzzer'.}
	\label{Fig:sound_demo}
\end{figure}

Fig.~\ref{Fig:sound_demo} illustrates the dynamics of single neurons that are trained with the tempotron (`Bin') and the TDP (`Mul') rule for a target class of `buzzer'. Neurons successfully elicit desired response to both target and null patterns as can be seen from the figure. With an imposed noise, neurons can still discriminate target patterns from null ones based on the output response, although the underlying dynamics is affected to a certain extent by the noise.

\begin{table}[!tbh]
	\centering
	\caption{Classification accuracy (in percentage \%) under mismatched condition. Shaded areas denote results obtained from our approaches in this study, while other baseline results are collected from \cite{dennis2013temporal,xiao2018spike,wu2018spiking}. The darker shaded color highlights our proposed multi-spike approach. The bold digits represent the best across each column. We use the `Avg$.$' as a performance indicator in the following evaluations, and the marker $\bullet$ is used in Fig.~\ref{Fig:sound_nd}-\ref{Fig:sound_ratio} for consistent presentation.}
	\begin{tabular}{c||ccccc|c}
		\hline
		\textit{Methods}  & Clean & 20dB & 10dB & 0dB & -5dB &  Avg$.$ \\ \hline 
		
		MFCC-HMM & 99.0  & 62.1  & 34.4 & 21.8  &  19.5 & 47.3 \\
		\rowcolor{LightCyan}
		SPEC-DNN    & \textbf{100}  & 94.38  & 71.8 & 42.68  &  34.85 & 68.74 \\
		\rowcolor{LightCyan}
		SPEC-CNN    & 99.83  & \textbf{99.88}  & 98.93 & 83.65  &  58.08 & 88.07 \\
		\rowcolor{LightCyan}
		KP-CNN    & 99.88  & 99.85  & \textbf{99.68} & 94.43  &  84.8 & 95.73 \\
		SOM-SNN & 99.6  & 79.15  & 36.25 & 26.5  &  19.55 & 52.21 \\
		
		LSF-SNN & 98.5  & 98.0  & 95.3 & 90.2  &  84.6 & 93.3 \\
		
		LTF-SNN & \textbf{100}  & 99.6  & 99.3 & 96.2  &  90.3 & 97.08 \\
		
		\rowcolor{LightCyan}
		KP-Bin   & 99.35  & 96.58  & 94.0 & 90.35  &  82.45 & 92.54 \\ 
		\rowcolor{DarkCyan}
		KP-Mul   & \textbf{100}  & 99.5  & 98.68 & \textbf{98.1}  &  \textbf{97.13} & \textbf{98.68} $\bullet$  \\		
		\hline
	\end{tabular}
	\label{tab:mismatch}
\end{table}

We examine the recognition performance under a mismatched condition where neurons are only trained with clean sounds but evaluated with different levels of noises after training. Table~\ref{tab:mismatch} shows the recognition performance of our proposed approaches, as well as other baseline ones.
Both conventional and spike-based approaches are covered. MFCC-HMM, as a typical framework widely used in acoustic processing, performs well in clean environment, but degrades rapidly with increasing levels of noise. The deep learning techniques, i.e. DNN and CNN, can be used to improve the performance as is compared to MFCC-HMM. CNN demonstrates a superior performance to DNN due to its advanced capabilities for processing images (here spectrograms, `SPEC') with convolution operations. Notably, the performance is significantly improved when we combine our key-point encoding (`KP') with CNN, reflecting the sparseness and effectiveness of our proposed encoding. 

On the other hand, the spike-based approaches try to solve the sound recognition task from a biologically realistic perspective. Most of these approaches except SOM-SNN get a relatively high performance with an average accuracy over 90\%. Different from other spike-based approaches \cite{dennis2013temporal,xiao2018spike,wu2018spiking} where a binary tempotron rule and a maximal voltage voting scheme are used, our proposed approach utilizes a more powerful multi-spike learning rule and adopts a simpler and more realistic maximal spike number voting scheme. Additionally, the comparative performance of our KP-Bin to that of LSF-SNN reflects the simplicity and efficiency of our encoding. In the following experiments, we use the average accuracy over all noisy conditions, i.e. `Avg$.$', as a performance index to further examine properties of our proposed system.

\begin{figure}[!htb]
	\centering\includegraphics[width=0.49\textwidth]{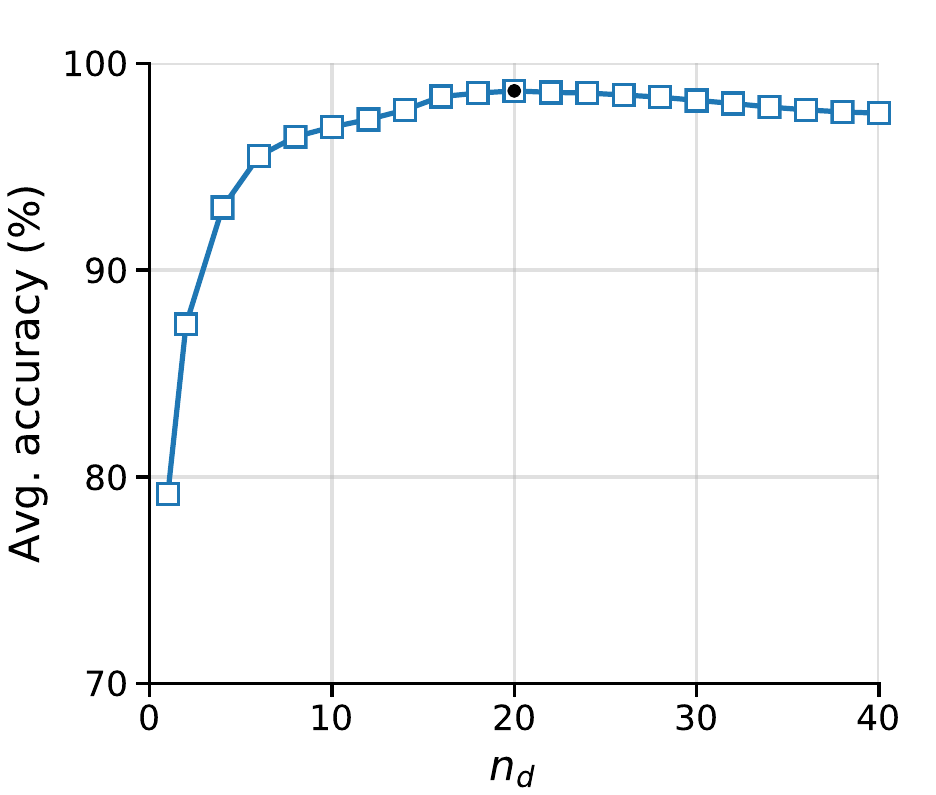}
	\caption{Recognition accuracy versus different target spike number $n_\mathrm{d}$.}
	\label{Fig:sound_nd}
\end{figure}

In Fig.~\ref{Fig:sound_nd}, we evaluate the effects of the target spike number on the performance of our multi-spike framework. In the case of $n_\mathrm{d}=1$, we restrict neurons to fire only single spikes in response to target classes. In this way, we can show the performance of a binary-spike decision scheme, and thus identify the contribution of a maximal voltage voting scheme applied in other approaches \cite{dennis2013temporal,xiao2018spike,wu2018spiking} and our KP-Bin in Table~\ref{tab:mismatch}. With an increasing number of $n_\mathrm{d}$, the performance of our approach is continuously enhanced. This is because more spikes can be used to explore useful features over time for a better decision. Further increasing $n_\mathrm{d}$ will marginally degrade the recognition accuracy which could be possibly due to a temporal interference \cite{yu2016spiking,rubin2010theory}.

\begin{figure}[!htb]
	\centering\includegraphics[width=0.49\textwidth]{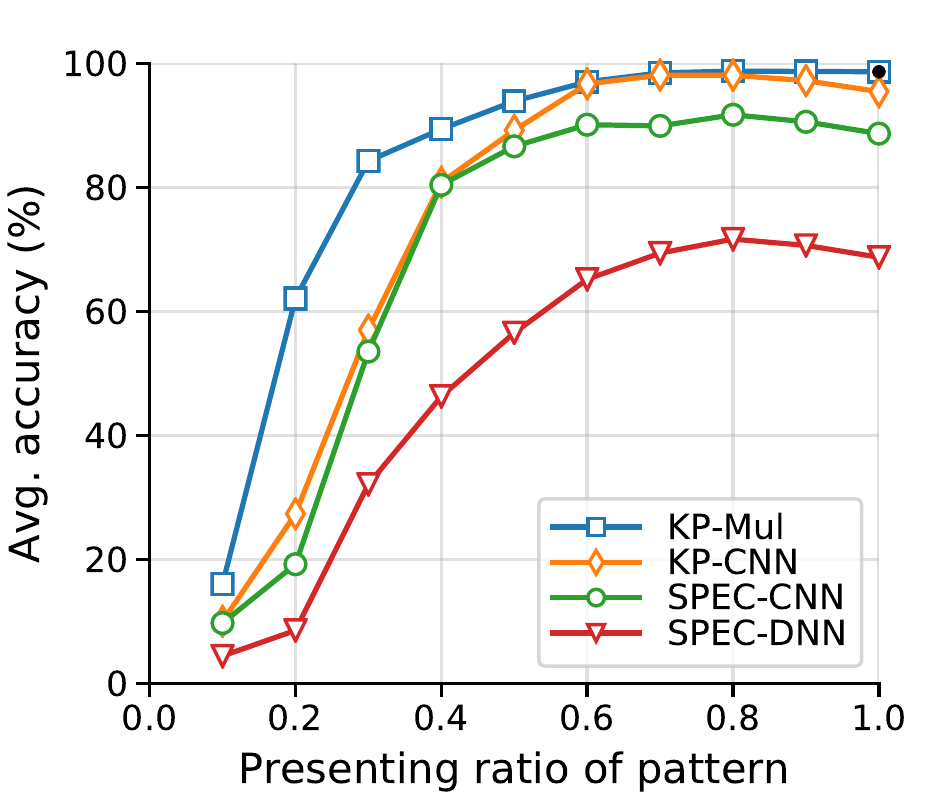}
	\caption{Recognition accuracy versus presenting ratio of single patterns.}
	\label{Fig:sound_early}
\end{figure}

Fig.~\ref{Fig:sound_early} shows the capabilities of different approaches to make early and prompt
discriminations when only a ratio of the whole pattern from the beginning is present to them.
As can be seen from the figure, different approaches favor longer presence of a pattern, because more useful information can be accumulated for a better decision. The larger the presenting ratio, usually the better the system performance. The performance of CNN and DNN based approaches slightly decreases when whole patterns are present. This is because the sound signals are recorded in a way with a short ending silence which does not contain useful information and thus distracts the deep learning approaches. This is not an issue to our multi-spike approach. The multi-spike characteristic makes neurons to elicit necessary spikes when more useful information appears and to ignore useless parts. Notably, our multi-spike approach demonstrates an outstanding capability for early decision as is compared to other ones. A relatively high accuracy of around 90\% can be obtained at an early time when only 40\% of the pattern is present.

\begin{figure}[!htb]
	\centering\includegraphics[width=0.49\textwidth]{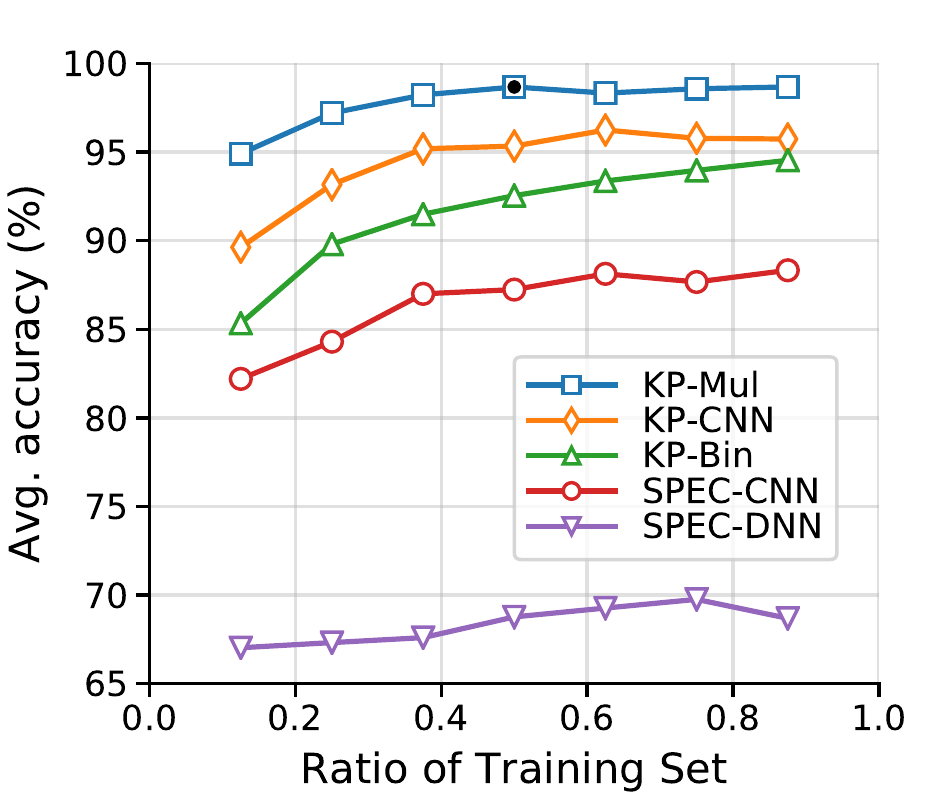}
	\caption{Recognition accuracy versus the ratio of the whole database used for training.}
	\label{Fig:sound_ratio}
\end{figure}

Fig.~\ref{Fig:sound_ratio} shows the effects of training-set sizes on the performance. All approaches generally demonstrate a higher accuracy for a larger training set. This is as expected since the more samples used for training the more knowledge neurons can learn. Importantly, our multi-spike approach achieves a high accuracy of around 95\% with only a small set used for training, indicating outstanding advantages of spike-based systems to learn from limited data samples which imposes difficulties for deep learning techniques \cite{lecun2015deep}.

In addition to the mismatched condition, we conduct multi-condition training to further improve the performance. The multi-condition training, which uses noise during learning, is found to be effective for performance enhancement \cite{lecun2015deep,YuQ2013PSD,wu2018spiking}. In our experiment, we randomly select conditions of clean, or noise levels of 20 or 10 dB during training. As can be seen from Table~\ref{tab:multicond}, the performance of our approaches is improved as expected. Our proposed multi-spike approach still outperforms other baseline ones.

\begin{table}[!tbh]
	\centering
	\caption{Classification accuracy (\%) under multi-condition training. The shaded area denotes the proposed multi-spike approach.}
	\begin{tabular}{c||ccccc|c}
		\hline
		\textit{Methods}  & Clean & 20dB & 10dB & 0dB & -5dB &  Avg$.$ \\ \hline 
			
		SPEC-DNN & 99.9  & 99.88  & 99.5 & 94.05  &  78.95 & 94.46 \\

		SPEC-CNN & 99.89  & 99.89  & 99.89 & 99.11  &  91.17 & 98.04 \\

		KP-CNN    & \textbf{99.93}  & 99.93  & 99.73 & 98.13 &  94.75 & 98.49 \\
		
		SOM-SNN	 & 99.8  & \textbf{100}  & \textbf{100} & \textbf{99.45} &  98.7 & \textbf{99.59} \\

		KP-Bin   & 99.13  & 99.23  & 99.1 & 95.1  &  89.38 & 96.38 \\ 
		\rowcolor{DarkCyan}
		KP-Mul   & 99.65  & 99.83  & 99.73 & 99.43  &  \textbf{98.95} & 99.52 \\		
		\hline
	\end{tabular}
	\label{tab:multicond}
\end{table}

\subsection{Processing Dynamic Sound Signals}

\begin{figure}[!htb]
	\centering\includegraphics[width=0.49\textwidth]{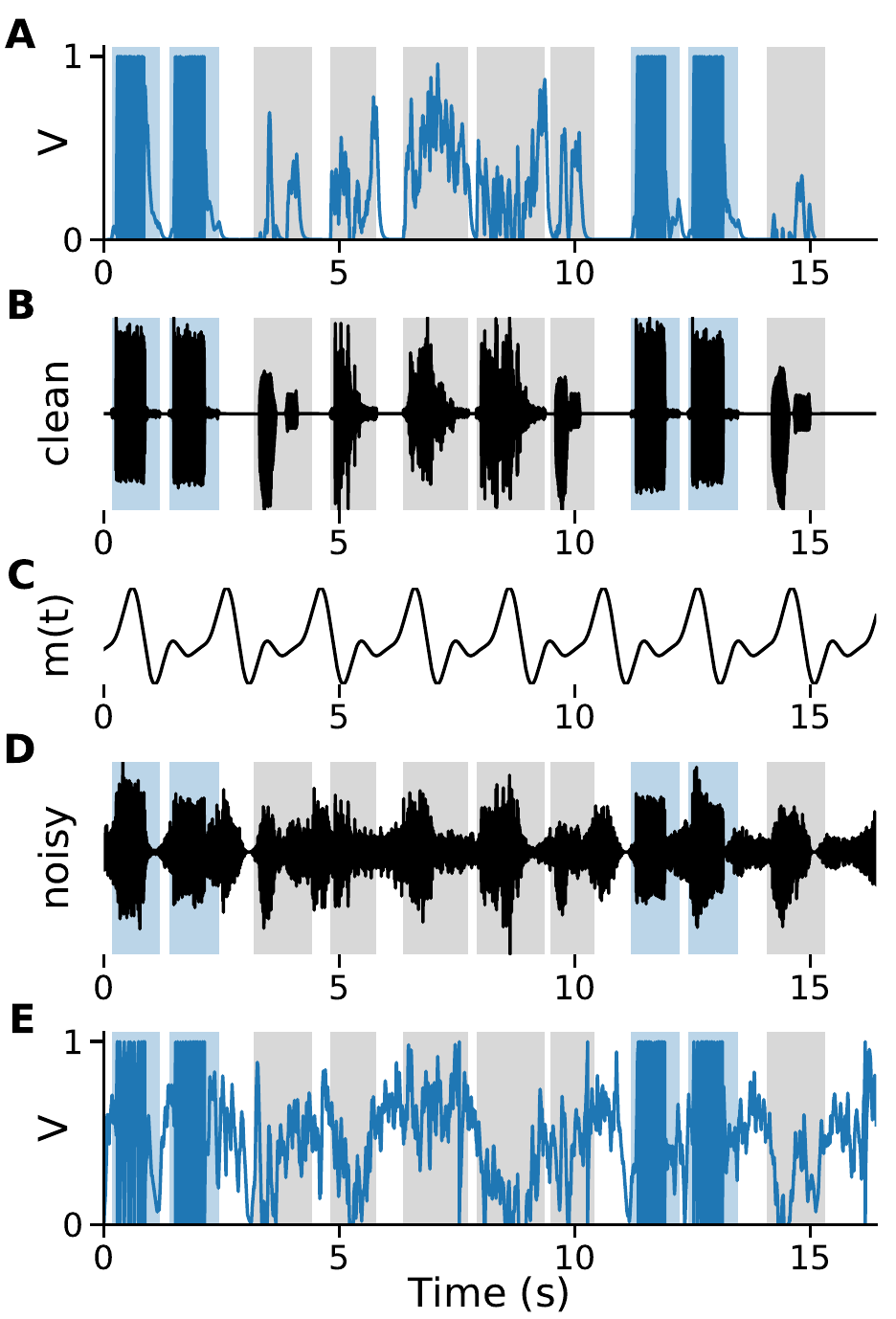}
	\caption{Demonstration of target sound extraction from ongoing dynamic environment. \textbf{A} and \textbf{E} are voltage traces of a neuron in response to sound signals under clean (\textbf{B}) and noisy (\textbf{D}) conditions, respectively. \textbf{C} is a modulator signal, $m(t)$, that is used in the noisy condition to simulate the highly time-varying characteristic of the noise. The blue and grey shaded areas denote the target and distractor sounds, respectively.}
	\label{Fig:sound_ongoing}
\end{figure}

In this experiment, we study the ability of our proposed framework for processing dynamic sound signals which is more challenging and realistic. In order to simulate the highly time-varying characteristic of a severe noise, we construct a modulator signal, $m(t)$, to modulate instantaneous power of the noise signal over time. The modulator is constructed as $m(t)=\sum_{i=1}^{3} A_i * \sin (2\pi f_i + \phi_i)$, where we set three frequencies ($f_i$) as 0.5, 1 and 1.5 Hz, and randomly choose the corresponding $A_i$ and $\phi_i$ from ranges of [0.0, 1.0] and [0, $2\pi$], respectively. Then, we linearly map $m(t)$ to the range of [0.0, 1.0] (see Fig.~\ref{Fig:sound_ongoing}\textbf{C} for an example). A strong SNR of -5 dB is used together with $m(t)$ to construct the noise signal. Both target and distractor sounds randomly occur over time.

As can be seen from Fig.~\ref{Fig:sound_ongoing}, the neuron can successfully detect and recognize the target sounds by a burst of firing spikes in both clean and noisy environment. 
Whenever a target sound appears, the neuron starts to continuously elicit spikes within the presenting duration of this target sound, while keeps silent for the other signals. The severe noise can significantly change the membrane potential dynamics of the neuron with a number of erroneous spikes being produced. However, a correction detection and recognition can still be made by the output spike density, i.e. bursting of spikes.

\section{Discussions}
\label{sec:discuss}

Our surrounding environment often contains both variant and non-stationary background noise. For example, a typical wind noise occasionally increases or decreases over time. Crucial sound events could occur arbitrarily in time and be embedded in the background noise. Successful recognition of such events is an important capability for both living individuals and artificial intelligent systems to adapt well in the environment. Inspired by the extraordinary performance of the brain on various cognitive tasks, we designed a biologically plausible framework for the environmental sound recognition task such that it can inherit various advantages such as efficiency and robustness from its counterparts in biology.

In our framework, spikes are used for both information transmission and processing, being an analogy of that in the central nervous systems \cite{kandel2000principles,dayan2001theoretical}. In addition to biological plausibility, spikes are believed to play an essential role in low-power consumption which would be of great importance to benefit devices such as mobiles and wearables where energy consumption is one of the major concerns \cite{merolla2014million}.
Moreover, spikes enable an event-driven computation mechanism which is more efficient as is compared to a clock-based paradigm \cite{yu2018spike}. These benefits drive efforts being devoted to delivering a paradigm shift toward more brain-like computations. Increasing number of neuromorphic hardwares have been developed in this direction recently with preliminary advantages being demonstrated \cite{merolla2014million,yao2017face,benjamin2014neurogrid,lichtsteiner2008128}.
We believe a synergy between neuromorphic hardwares and systems like ours could push the spike-based paradigm to a new horizon.

Our framework is a unifying system that consists of three major functional parts including encoding, learning and readout (see Fig.~\ref{Fig:framework}). All the three parts are consistently integrated in a same spike-based form. In our encoding frontend, key-points are detected by localizing the sparse high-energy peaks in the spectrogram. These peaks are inherently robust to mismatched noise due to the property of local maximum, and are sufficient enough to give a `glimpse' of the sound signal with a sparse representation (see Fig.~\ref{Fig:kp}). The sparseness and robustness of our encoding can significantly increase the performance even of a conventional classifier (see `KP-CNN' in Table~\ref{tab:mismatch}), indicating that its effectiveness could be generalized. Importantly, our encoding does not rely on any auxiliary networks such as SOM used in \cite{dennis2013temporal,wu2018spiking}, or frame-based time reference \cite{wu2018spiking}, making our frontend a superior choice for on-the-fly processing due to this simplicity.

An efficient multi-spike learning rule, i.e. TDP, is employed in our learning part. The TDP is recently proposed to train neurons with a desired number of spikes \cite{yu2018spike}. The learning efficiency gives priority to TDP over other multi-spike learning rules such as MST and PSD (see Fig.~\ref{Fig:efficiency}). Additionally, TDP can automatically explore and exploit more temporal information over time without specifically instructing neurons at what time to fire. 
A more robust and better performance is thus obtained (see Fig.~\ref{Fig:robustness}). The TDP rule is capable of processing not only temporal encoded spike patterns but also rate-based ones \cite{yu2018spike}. Fig.~\ref{Fig:rt} demonstrates the capability of TDP to extract information from inhomogeneous firing statistics which is more challenging as is compared to a homogeneous one. All of our property examinations on the learning rules suggest TDP as a perfect choice for sound recognition. Moreover, these examinations could provide a useful reference for spike-based developments.

In our readout part, we utilize the maximal output spike number to indicate the category of an input pattern. The more spikes a neuron elicits, the higher likelihood the presenting pattern belongs to the neuron's category. With this multi-spike readout, the performance of sound recognition is significantly improved as is compared to a single-spike one (see Fig.~\ref{Fig:sound_nd}). The neuron can make full use of the temporal structure of sound signals over time by firing necessary spikes in response to useful information (see Fig.~\ref{Fig:sound_demo} for a demonstration). An impressive early-decision making property is also a result of this readout scheme (see Fig.~\ref{Fig:sound_early}). This behavior highlights the advantages of brain-inspired SNNs to respond with only a few early spikes. Notably, the performance can be improved by further accumulating useful information that occurs after early times (see Fig.~\ref{Fig:sound_early}). The performance of a binary-spike learning rule can be enhanced by a maximal potential readout \cite{dennis2013temporal,wu2018spiking,xiao2018spike}, but this scheme is inefficient since a maximal value needs to be tracked over time. Differently, our multi-spike readout is as simple as to count spike appearance only, thus benefiting both software and hardware implementations.

The spike-based framework is naturally suitable for processing temporal signals. The outstanding performance of our spike-based approaches (see Table~\ref{tab:mismatch} and \ref{tab:multicond}) over the conventional baseline methods highlights the computational power of brain-like systems. 
The capability of our framework to process ongoing signal streams in an efficient and robust way endows it with great advantages for practical applications such as bioacoustic monitoring \cite{weninger2011audio}, 
surveillance \cite{ntalampiras2009acoustic} and general machine hearing \cite{lyon2010machine}.

\section{Conclusion}
\label{sec:Conclusion}
In this work, we proposed a spike-based framework for environmental sound recognition. The whole framework was consistently integrated together with functional parts of sparse encoding, efficient learning and robust readout. Firstly, we introduced a simple and sparse encoding frontend where key-points are used to represent acoustic features. It was shown that our encoding can be generalized to even benefit other non-spike based methods such as DNN and CNN, in addition to our spike-based systems. Then, we examined properties of various spike learning rules in details with our contributions being added for improvements. We showed that the adopted multi-spike learning rule is efficient in learning and powerful in processing spike streams without restricting a specific spike coding scheme. Our evaluations could be instructive not only to the selection of rules in our task but also to other spike-based developments.
Finally, we combined a multi-spike readout with the other parts to form a unifying framework. We showed that our system performs the best in the mismatched sound recognition as is compared to other spike or non-spike based approaches. Performance can be further improved by adopting a multi-condition training scheme. Additionally, our framework was shown to have various advantageous characteristics including early decision making, small dataset training and ongoing dynamic processing.
To the best of our knowledge, our framework is the first work to apply the multi-spike characteristic of nervous neurons to practical sound recognition tasks. The preliminary success of our system would pave the way for more research efforts to be made to the neuromorphic, i.e. spike-based, domain.

%


\ifCLASSOPTIONcaptionsoff
  \newpage
\fi

\end{document}